\documentclass[times,twocolumn,final]{elsarticle}

\usepackage{framed,multirow}

\usepackage{amssymb}
\usepackage{latexsym}

\usepackage{url}
\usepackage{xcolor}

\usepackage{pifont}
\newcommand{\cmark}{\ding{51}}%
\newcommand{\xmark}{\ding{55}}%

\usepackage[hidelinks]{hyperref}
\definecolor{newcolor}{rgb}{.8,.349,.1}

\usepackage[ruled,vlined,algo2e]{algorithm2e}
\SetKwComment{Comment}{$\triangleright$\ }{}

\usepackage{mathtools}
\usepackage{amsmath}

\newcommand\Tstrut{\rule{0pt}{2.0ex}}         
\newcommand\Bstrut{\rule[-0.9ex]{0pt}{0pt}}   

\begin{document}


\begin{frontmatter}

\title{2.5D Multi-view Averaging Diffusion Model for 3D Medical Image Translation: Application to Low-count PET Reconstruction with CT-less Attenuation Correction}

\author[2,4]{Tianqi Chen}
\author[2,4]{Jun Hou}
\author[1]{Yinchi Zhou}
\author[1]{Huidong Xie}
\author[1]{Xiongchao Chen}
\author[1]{Qiong Liu}
\author[1]{Xueqi Guo}
\author[2]{Menghua Xia}
\author[1,2,3]{James S. Duncan}
\author[1,2]{Chi Liu}
\author[1,5]{Bo Zhou\corref{cor1}}
\cortext[cor1]{Corresponding author.}
\ead{bo.zhou@northwestern.edu}

\address[1]{Department of Biomedical Engineering, Yale University, New Haven, CT, USA}
\address[2]{Department of Radiology and Biomedical Imaging, Yale School of Medicine, New Haven, CT, USA}
\address[3]{Department of Electrical Engineering, Yale University, New Haven, CT, USA.}
\address[4]{Department of Computer Science, University of California Irvine, Irvine, CA, USA}
\address[5]{Department of Radiology, Northwestern University, Chicago, IL, USA.}





\begin{abstract}
Positron Emission Tomography (PET) is an important clinical imaging tool but inevitably introduces radiation hazards to patients and healthcare providers. Reducing the tracer injection dose and eliminating the CT acquisition for attenuation correction can reduce the overall radiation dose, but often results in PET with high noise and bias. Thus, it is desirable to develop 3D methods to translate the non-attenuation-corrected low-dose PET (NAC-LDPET) into attenuation-corrected standard-dose PET (AC-SDPET). Recently, diffusion models have emerged as a new state-of-the-art deep learning method for image-to-image translation, better than traditional CNN-based methods. However, due to the high computation cost and memory burden, it is largely limited to 2D applications. To address these challenges, we developed a novel 2.5D Multi-view Averaging Diffusion Model (MADM) for 3D image-to-image translation with application on NAC-LDPET to AC-SDPET translation. Specifically, MADM employs separate diffusion models for axial, coronal, and sagittal views, whose outputs are averaged in each sampling step to ensure the 3D generation quality from multiple views. To accelerate the 3D sampling process, we also proposed a strategy to use the CNN-based 3D generation as a prior for the diffusion model. Our experimental results on human patient studies suggested that MADM can generate high-quality 3D translation images, outperforming previous CNN-based and Diffusion-based baseline methods. 

\end{abstract}


\end{frontmatter}


\section{Introduction}
\begin{figure}[htb!]
\centering
\includegraphics[width=0.356\textwidth]{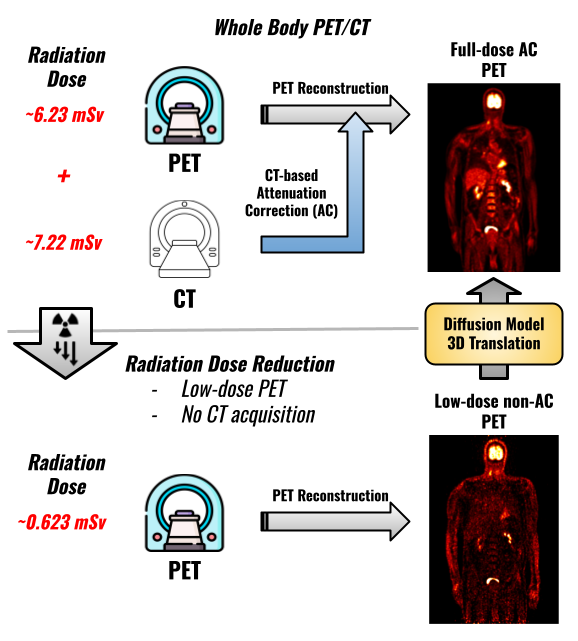}
\caption{Illustration of low-count/dose PET reconstruction with CT-less Attenuation Correction (AC) for reducing the overall radiation dose in PET.}
\label{fig:intro}
\end{figure}

Positron emission tomography (PET) is a vital functional imaging tool with wide applications in oncology, cardiology, neurology, and biomedical research. In clinical practice, PET imaging involves radioactive tracer injection to the patients, with the dosage level carefully guided by the principle of As Low As Reasonably Achievable (ALARA) \cite{strauss2006alara}. In order to correct the attenuation of the PET signal, PET is also acquired along with CT to provide an attenuation map ($\mu$-map) for PET attenuation correction (AC). This multi-modal imaging procedure inevitably introduces radiation hazards to the patient and healthcare providers. Thus, it is highly desirable to reduce the overall radiation dose in this important imaging procedure. On one hand, the PET radiation dose can be directly reduced by reducing the tracer injection dose. However, the PET image will suffer from high noise and bias with the low-count signal. On the other hand, while PET can be acquired alone without CT, the PET image would inevitably suffer further increased quantification errors if CT-based AC is not included. Thus, reconstruction of standard-count PET from low-count PET and CT-less AC for PET are both important topics for reducing the overall radiation dose in PET imaging, as illustrated in Figure \ref{fig:intro}. 

There are many previous works on low-count PET image denoising, and can be summarized into two categories, including conventional image post-processing \cite{dutta2013non,maggioni2012nonlocal,mejia2016noise} and deep learning (DL) based methods \cite{xiang2017deep,wang20183d,lu2019investigation,kaplan2019full,hu2020dpir,gong2020parameter,zhou2020supervised,ouyang2019ultra,chen2019ultra,liu2020noise,xie_ddpet-3d_2023,pan2024full}. Conventional image post-processing techniques, such as Gaussian filtering, are standard techniques in PET reconstruction but hard to preserve local structures. Non-local mean filter \cite{dutta2013non} and block-matching 4D filter \cite{maggioni2012nonlocal} were proposed to denoise low-dose PET while better preserving the structural information. While these conventional image post-processing methods may substantially improve the image quality, over-smoothing is commonly observed in ultra-low-count settings. As the statistical characteristics of noise in medical imaging are complex and hard to model, DL models can learn the highly non-linear relationship from data and recover the original signal from noise. Previous DL-based methods can be further divided into two categories. The first category only uses the low-count PET data as input. For example, Kaplan et al. \cite{kaplan2019full} proposed using a GAN \cite{goodfellow2014generative} with UNet \cite{ronneberger2015u} as the generator to predict standard-count PET from low-count PET. Similarly, Wang et al. \cite{wang20183d} proposed using a 3D-conditional-GAN \cite{isola2017image} also with UNet as the generator to translate low-count PET to standard-count PET. Zhou et al. \cite{zhou2020supervised} and Gong et al. \cite{gong2020parameter} found incorporating Wasserstein GAN \cite{arjovsky2017wasserstein} can also achieve promising low-count PET denoising performance. The second category uses low-count PET and MR/CT as input. For example, Xiang et al. \cite{xiang2017deep} proposed a deep auto-context CNN that takes low-count PET image and T1 MR image as input for prediction of standard-count PET. Similarly, Chen et al. \cite{chen2019ultra} proposed to input low-count PET along with multi-contrast MR images into a UNet \cite{ronneberger2015u} for ultra-low-dose PET denoising. Compared to conventional PET denoising methods, all these DL-based methods achieved superior denoising performance. However, almost all of these previous methods were developed based on U-Net or its variants\cite{ronneberger2015u,isola2017image,zhu2017unpaired}. Most importantly, they assumed CT-based AC is already done for the low-count PET, thus CT-less AC for further reducing the radiation was not considered.  

\begin{figure*}[htb!]
\centering
\includegraphics[width=0.946\textwidth]{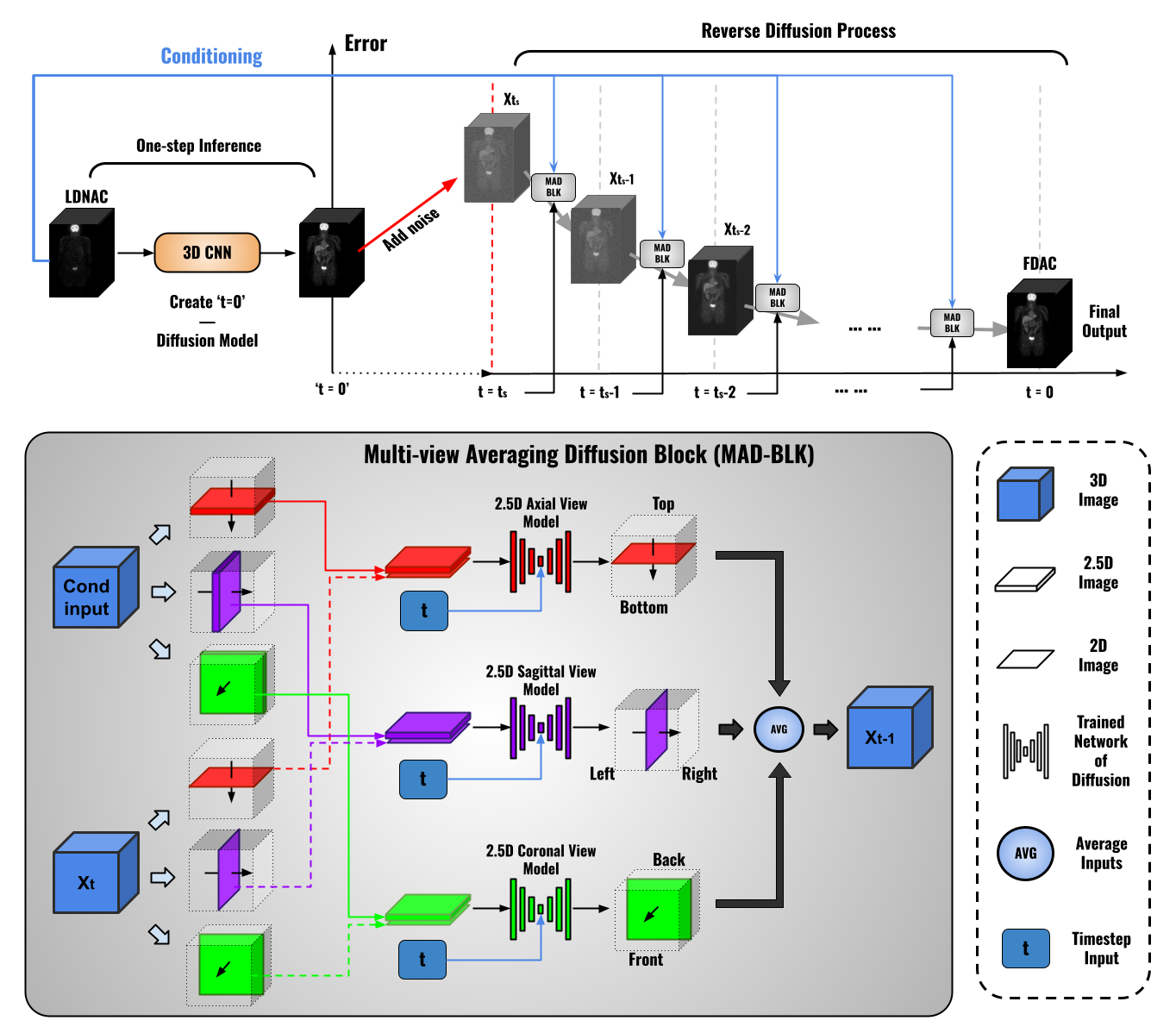}
\caption{The overall workflow of 2.5D \textbf{M}ulti-view \textbf{A}veraging \textbf{D}iffusion \textbf{M}odel (MADM). MADM contains a one-step inference generative model(orange) and MAD-BLK(grey). The MAD-BLK contains Three models in axial, sagittal, and coronal view, and the output of each model will be averaged in the average block before output.}
\label{fig:pipeline}
\end{figure*}

\begin{algorithm2e*}[htb!]
\label{alg:MADM}
\(\textbf{Input: } x\in N^{d_1\times d_2\times d_3}\)\\
\(\textbf{Initialize \#1: }t_s\in[0, T]\text{: the start timestep of denoising process}\)\\
\(\textbf{Initialize \#2: }s\text{: the number of adjacent slices loaded as 2.5D condition input}\)\\
\(\textbf{Initialize \#3: }\epsilon \sim N(0, I) \)\\
\(y_{\text{prior}} = f_{\text{prior}}(x)\) \Comment*[r]{One-step inference to generate prior image}
\(\bar{y}_{\text{prior}} = \sqrt{\bar{a}_{t_0}} y_{\text{prior}} + \sqrt{1 - \bar{a}_{t_0}} \epsilon\) \Comment*[r]{Diffuse, i.e. add noise, to the prior image}
\For{$t = t_s, t_{s}-1, ..., 0$}{ 
    \(y^{\text{avg}}_{t} = \bar{y}_{\text{prior}}\text{ if }t == t_s\) \Comment*[r]{Start the reverse process at $t_s$ based on diffused prior image}
    \(z \sim \mathcal{N}(0, \mathcal{I})\text{ if }t > 1\text{, else }z = 0\)\\
    \For{$i = 1, 2, ..., d_1$}{
        \(y^{\text{co}}_{t-1}[i,:,:] = \frac{1}{\sqrt{\alpha_t}}(y^{\text{avg}}_t[i,:,:] - \frac{1-\alpha_t}{\sqrt{1-\bar{\alpha_t}}}f^{co}_{\theta}(y^{\text{avg}}_t[i,:,:], x[i-s:i+s,:,:], t)) + \sigma_t z\) \Comment*[r]{Coronal view 2.5D reverse}
    }
    \For{$i = 1, 2, ..., d_2$}{
        \(y^{\text{sa}}_{t-1}[:,i,:] = \frac{1}{\sqrt{\alpha_t}}(y^{\text{avg}}_t[:,i,:] - \frac{1-\alpha_t}{\sqrt{1-\bar{\alpha_t}}}f^{sa}_{\theta}(y^{\text{avg}}_t[:,i,:], x[:,i-s:i+s,:], t)) + \sigma_t z\) \Comment*[r]{Sagittal view 2.5D reverse}
    }
    \For{$i = 1, 2, ..., d_3$}{
        \(y^{\text{ax}}_{t-1}[:,:,i] = \frac{1}{\sqrt{\alpha_t}}(y^{\text{avg}}_t[:,:,i] - \frac{1-\alpha_t}{\sqrt{1-\bar{\alpha_t}}}f^{ax}_{\theta}(y^{\text{avg}}_t[:,:,i], x[:,:,i-s:i+s], t)) + \sigma_t z\) \Comment*[r]{Axial view 2.5D reverse}
    }
    \(y^{\text{avg}}_{t-1} = \frac{y^{\text{co}}_{t-1} + y^{\text{ax}}_{t-1} + y^{\text{sa}}_{t-1}}{3}\) \Comment*[r]{Averaging volumes from multiple 2.5D view models}
}
\end{algorithm2e*}

On the other hand, there are also many previous efforts for CT-less AC for PET. For time-of-flight (TOF) PET imaging data, Maximum Likelihood Reconstruction of Activity and Attenuation (MLAA) algorithms \cite{rezaei2016simultaneous} were first developed to simultaneously reconstruct the tracer activity ($\lambda$-MLAA) and the attenuation map ($\mu$-MLAA), based on the TOF PET raw data only. Ideally, MLAA would solve the AC problem for TOF PET. However, the direct use of $\mu$-MLAA still suffers from significant quantification error as compared to the CT-based ($\mu$-CT) OSEM reconstruction \cite{rezaei2016simultaneous,shi2023deep,nuyts2018validation}, even in the standard-count PET. With recent advancements in deep learning (DL), DL-based CT-less AC methods have been extensively investigated to address this challenge \cite{lee2020review,chen2023deep}. The previous DL-based AC methods can be summarized into two classes. The first category of methods uses either non-attenuation corrected (NAC) PET or MLAA reconstructions as deep network input to generate CT-based $\mu$-map \cite{shi2023deep,toyonaga2022deep,liu2018deep1,liu2018deep2,hwang2019generation,dong2019synthetic}. Then, AC is performed based on the generated $\mu$-map. Similarly, the second category of methods also uses either NAC PET or MLAA reconstructions as deep network input, but to directly generate attenuation-corrected PET \cite{shiri2019direct,dong2020deep,shiri2023differential}. However, almost all of these previous methods were developed based on U-Net or its variants\cite{ronneberger2015u,isola2017image,zhu2017unpaired} and only validated on standard-count PET. Even though these approaches achieved reasonable AC performance on standard-count PET, directly deploying these methods for low-count PET AC may result in non-ideal performance given the low-quality PET input. Additional steps for low-count PET denoising also need to be considered even after AC is performed.

While reconstruction of standard-count PET from low-count PET task and CT-less AC for standard-count PET task were both extensively studied in prior works, low-count PET reconstruction with CT-less AC that can holistically reduce radiation dose has not been investigated before. Directly deploying previous DL models may lead to non-ideal translation results since the low-count PET image without AC not only suffers from high noise levels but also significant bias due to attenuation. In recent years, the Diffusion Model (DM) has emerged as the new state-of-the-art image generation method and has shown promising results in image-to-image translation. However, there are two major issues with DM. First, these methods were mainly developed for 2D image data, but not 3D data like PET. Translating 3D images in a slice-by-slice 2D manner will lead to inconsistency from another view \cite{xie_ddpet-3d_2023, lee2023improving, chung_solving_2023}. Even though it is possible to directly extend the 2D network into the 3D network in the diffusion model for 3D translation, the computation burden and memory requirement make it highly infeasible with the current hardware technique, especially when it is applied to high-resolution 3D imaging data like PET. Second, the translation speed is significantly slower than previous CNN-based approaches, due to the large number of diffusion sampling steps required. This is further exacerbated if DM is directly extended to 3D imaging data like PET. 

To address these issues, we developed a novel 2.5D Multi-view Averaging Diffusion Model (MADM) for 3D medical image translation with the first application on low-count PET reconstruction with CT-less AC. There are three main contributions of MADM. First, to reduce the computational demands associated with 3D models and to mitigate the discontinuities inherent in 2D models, we proposed to use a 2.5D model that incorporates several adjacent slices in addition to the target slice for prediction. Second, to further minimize discontinuities and enhance prediction accuracy, we proposed to train separate 2.5D models for each view in 3D, i.e. axial, coronal, and sagittal views. During the inference, we average the result in each denoising step of the diffusion process. Third, to expedite the translation process and refine generated image accuracy, we proposed to utilize 3D predictions from a Convolutional Neural Network (CNN)-based model as a prior for our diffusion model. Our experimental results show that MADM can generate high-quality AC standard-count PET(AC-SDPET) directly from the NAC low-count PET(NAC-LDPET) in 3D with reasonable computation resources, and also surpassing several previous baselines. 


\section{Methods}
\subsection{Cascaded Multi-path Shortcut Diffusion Model}
The general pipeline of the 2.5D \textbf{M}ulti-view \textbf{A}veraging \textbf{D}iffusion \textbf{M}odel (MADM) is depicted in Figure \ref{fig:pipeline}. During both training and sampling processes, MADM employs a multifaceted approach: first, we use a CNN-based one-step inference generative model to create a 3D prior image, then we use multiple conditional 2.5D diffusion models in axial, coronal, and sagittal view and average the outputs from different views in each denoising step to further refine the 3D prior image. The following section details the training and inference methodologies.

\subsection{2.5D Multi-view Averaging Diffusion} 
We denote the NAC-LDPET input image as $x\in N^{d_1\times d_2\times d_3}$ and the target AC-SDPET image as $y\in N^{d_1\times d_2\times d_3}$ where $d_1, d_2, d_3$ is the width, depth, and hight of the 3D image. 
\\[6pt]
\noindent\textbf{Training:} To generate the prior image for MADM, we first need to train a one-step inference generative model $f_\text{prior}(\cdot)$, i.e. U-Net \cite{ronneberger2015u} here, which predicts $y_0$ from $x$. This model can be trained using the L2 loss
\begin{equation} \label{eq:priorloss}
    \mathcal{L}_{v} = ||f_{\text{prior}} (x) - y||^2_2,
\end{equation}

On the other hand, the diffusion process tries to predict the distribution $y_0\sim q(y)$. Then we have a forward diffusion process which adds noise $\epsilon$ to $y_0$ for $T$ steps:
\begin{equation} \label{eq:forward}
    q(y_t|y_{t-1}) = \mathcal{N}(y_t;\sqrt{1-\beta_t}y_{t-1}, \beta_tI),
\end{equation}
and 
\begin{equation} \label{eq:forward1T}
    q(y_{1:T}|y_{0}) = \prod^T_{t=1}q(y_t|y_{t-1}),
\end{equation}
where we let ${\beta_t \in (0,1)}^T_{t=1}$ and $\beta_t < \beta_{t+1}$. Following that, we can sample $y_t$ for any $t \in [0, T]$ using:
\begin{equation} \label{eq:addnoise}
    y_t = \sqrt{\bar{\alpha}_t}y_0 + \sqrt{1 - \bar{\alpha}_t}\epsilon,
\end{equation}
\begin{equation} \label{eq:forwardqt}
    q(y_t|y_0) = \mathcal{N}(y_t;\sqrt{\bar{\alpha}_t}y_0, (1-\bar{\alpha}_t) I),
\end{equation}
where $\epsilon \sim N(0, I)$, $\alpha_t = 1 - \beta_t$, and $\bar{\alpha_t} = \prod^t_{i=1}\alpha_i$. Then in the reverse diffusion process(denoising process), we want to sample $y_0$ from $y_T \sim \mathcal{N}(0, I)$, thus we need a model $p_{\theta}$ to run this reverse diffusion process which
\begin{equation} \label{eq:backward}
    p_\theta(y_{t-1}|y_t) = \mathcal{N}(y_{t-1};\mu_{\theta}(y_t, x, t), \beta_tI),
\end{equation}
and 
\begin{equation} \label{eq:backward1T}
    p_{\theta}(y_{0:T}) = p(y_T)\prod^T_{t=1}p_{\theta}(y_{t-1}|y_{t}),
\end{equation}
where 
\begin{equation} \label{eq:predmu}
    \mu_{\theta}(y_t, x, t) = \frac{1}{\sqrt{\alpha_t}}(y_t - \frac{1-\alpha_t}{\sqrt{1-\bar{\alpha}_t}}f_\theta(y_t, x, t)),
\end{equation}

Therefore we can use model $f_\theta(\cdot)$ to predict $y_{t-1}$ from $y_t$ given the NAC-LDPET input $x$.
In our MADM, instead of using a whole 3D $x$ as the conditional input, we use 2.5D slices $x^v$ cropped $(2s+1)$ adjacent slices from $x$ in different views $v \in \{\text{coronal}, \text{sagittal}, \text{axial}\}$.

To train the conditional diffusion model $f^v_{\theta}(\cdot)$ separately in coronal, sagittal, and axial view, we use pixel-wise L2 loss
\begin{equation} \label{eq:loss}
    \mathcal{L}_{v} = ||f_{\theta}^{v} (y_t^v, x^v, t) - \epsilon||^2_2,
\end{equation}
where $t = 0, 1, ..., T$ is the timestep in diffusion process.
\\[6pt]
\noindent\textbf{Sampling:}After the loss of conditional diffusion models $f_{\theta}^{v}$ and $f_{\text{prior}}$ converged in training, we use $f_{\text{prior}}$ to predict the $y_{\text{prior}}$ from $x$
\begin{equation} \label{eq:prior}
    y_{\text{prior}} = f_{\text{prior}}(x).
\end{equation}
Then we add noise to the $y_\text{{prior}}$ to the same noise level $t_s$ in diffusion process by
\begin{equation} \label{eq:add_noise_to_prior}
    \bar{y}_{\text{prior}} = \sqrt{\bar{a}_{t_s}} y_{\text{prior}} + \sqrt{1 - \bar{a}_{t_s}} \epsilon,
\end{equation}
where $t_s \in [0, T]$. Following the DDPM process, the noisy image in $t-1$ for the step $t$ can be predicted using the conditional diffusion models
\begin{equation} \label{eq:predict}
    y^{\text{v}}_{t-1} = \frac{1}{\sqrt{\alpha_t}}(y^v_t - \frac{1-\alpha_t}{\sqrt{1-\bar{\alpha_t}}}f^v_{\theta}(y^v_t, x^v, t)) + \sigma_t z,
\end{equation}
where $y_{t_s} = \bar{y}_{\text{prior}}$ when $t = t_s$. After predicting the $y_{t-1}^v$ in three views, our method averages the three predictions using
\begin{equation} \label{eq:average}
    y^{\text{avg}}_{t-1} = \frac{y^{\text{co}}_{t-1} + y^{\text{ax}}_{t-1} + y^{\text{sa}}_{t-1}}{3},
\end{equation}
to get the final prediction for step $t-1$ then repeat the denoise process using equation \ref{eq:predict} and \ref{eq:average} from step $t_s$ to $0$ to get the final result $y_0 = y_{0}^{avg}$. The algorithm is summarized in Algorithm \ref{alg:MADM}.






\begin{figure*}[htb!]
\centering
\includegraphics[width=0.98\textwidth]{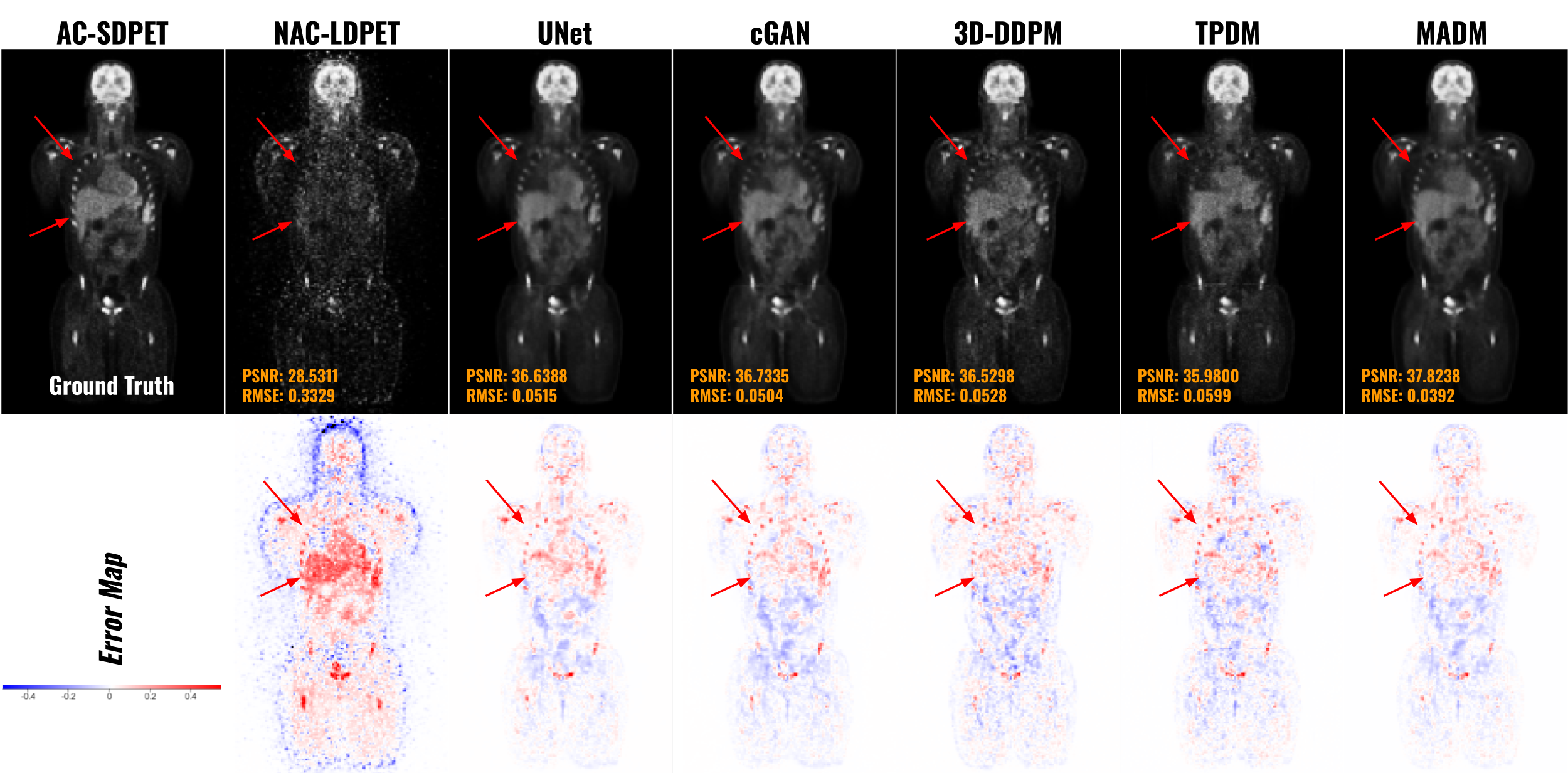}
\caption{Visual comparison of AC-SDPET generation from different methods under 5\% NAC-LDPET settings. The coronal view image(top) and error map(bottom) are shown. RMSE and PSNR values are calculated for each individual volume. The CT-based AC-SDPET(top) and image index in to the color map of the error map(bottom) are shown in the first column.}
\label{fig:comp_methods}
\end{figure*}
\begin{figure*}[htb!]
\centering
\includegraphics[width=0.98\textwidth]{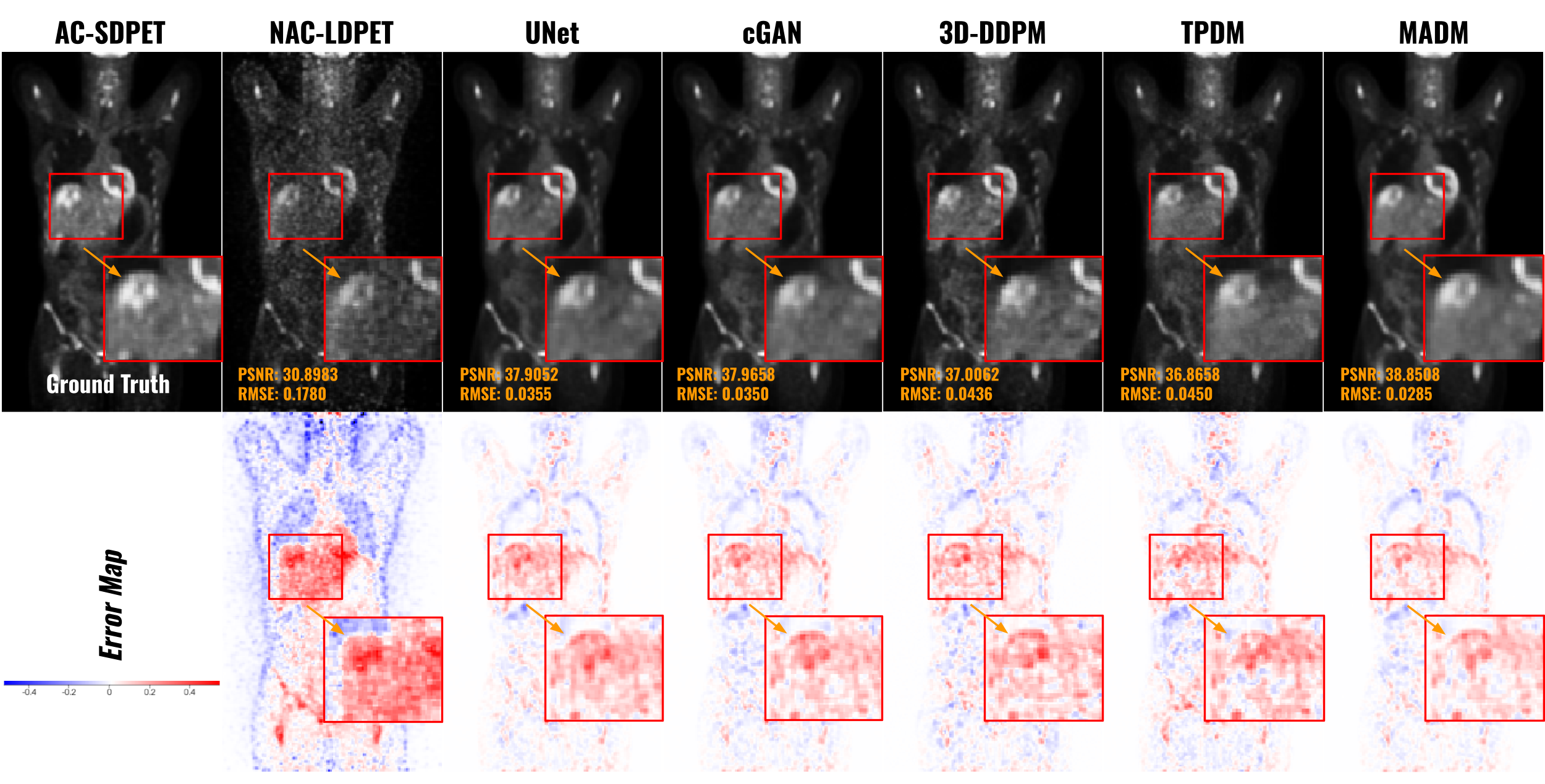}
\caption{Visual comparison of AC-SDPET generation from different methods under 10\% NAC-LDPET settings. The coronal view image(top) and error map(bottom) are shown. RMSE and PSNR values are calculated for each individual volume. The CT-based AC-SDPET(top) and image index in to the color map of the error map(bottom) are shown in the first column.}
\label{fig:comp_methods_10NAC}
\end{figure*}
\begin{table*} [htb!]
\footnotesize
\centering
\caption{Quantitative comparison of AC-SDPET generation from different methods under two different low-count settings using PSNR, SSIM, and RMSE. Best results are marked in \textbf{bold}.}
\label{tab:comp_methods}
\resizebox{0.98\textwidth}{!}{
    \begin{tabular}{l|c|c|c||c|c|c||c|c}
        \hline
        \multirow{2}{*}{\textbf{Evaluation}}  & \multicolumn{3}{c}{\textbf{5\% Low-count}}                   &  \multicolumn{3}{c}{\textbf{10\% Low-count}}     &  \multicolumn{2}{c}{\textbf{GPU Memory Usage}}               \Tstrut\Bstrut\\
        \cline{2-9}
                                              & PSNR             & SSIM              & RMSE              & PSNR            & SSIM              & RMSE       &Training(MB) &Sampling(MB)         \Tstrut\Bstrut\\
        \hline
        NAC-LD                    & $29.385 \pm 1.440$   & $0.9489 \pm 0.0147$     & $0.2666 \pm 0.0621$     & $30.049 \pm 1.442$   & $0.9544 \pm 0.0133$  & $0.2293 \pm 0.0561$  & - & -    \Tstrut\Bstrut\\
        \hline
        UNet\cite{10.1007/978-3-319-24574-4_28}               & $38.101 \pm 1.423$   & $0.9936 \pm 0.0021$     & $0.0360 \pm 0.0096$     & $38.869 \pm 1.384$   & $0.9946 \pm 0.0018$  & $0.0301 \pm 0.0082$    &10014 &10460  \Tstrut\Bstrut\\
        \hline
        cGAN\cite{Isola_2017_CVPR}               & $38.295 \pm 1.467$   & $0.9938 \pm 0.0021$     & $0.0344 \pm 0.0087$     & $39.045 \pm 1.454$   & $0.9948 \pm 0.0018$  & $0.0289 \pm 0.0076$  &11440 &13020     \Tstrut\Bstrut\\
        \hline
        3D-DDPM\cite{bieder2023diffusion}            & $37.197 \pm 1.271$   & $0.9924 \pm 0.0026$     & $0.0437 \pm 0.0085$     & $37.541 \pm 1.189$   & $0.9930 \pm 0.0022$  & $0.0403 \pm 0.0079$  &68874 &23798     \Tstrut\Bstrut\\
        \hline
        TPDM\cite{lee2023improving}               & $36.818 \pm 1.461$   & $0.9916 \pm 0.0029$     & $0.0485 \pm 0.0150$     & $37.463 \pm 1.478$   & $0.9925 \pm 0.0027$  & $0.0419\pm 0.0137$    &6442 &3012   \Tstrut\Bstrut\\
        \hline
        \hline
        MADM                      & $\mathbf{38.920 \pm 1.503}$   & $\mathbf{0.9944 \pm 0.0020}$    & $\mathbf{0.0299 \pm 0.0088}$    & $\mathbf{39.453 \pm 1.487}$   & $\mathbf{
        0.9952\pm 0.0017}$    & $\mathbf{0.0265 \pm 0.0083}$   &6442 &4464   \Tstrut\Bstrut\\
        \hline
    \end{tabular}
}
\end{table*}

\subsection{Data Preparation and Implementation Details}
We collected a real patient dataset from the Yale New Haven Hospital (YNHH) for training and evaluation of our method. The data was acquired using a Siemens Biograph mCT scanner at YNHH. PET scans were performed approximately 60 mins after intravenous injection of about 10 mCi \textsuperscript{18}F-FDG tracer, with whole-body continuous bed scanning protocol. We used uniform down-sampling of the PET list-mode data with down-sampling ratios of 5\% and 10\% to generate low-count PET data at two different low-count levels. To generate the non-AC low-count PET image (i.e. NAC-LDPET), we used the down-sampled PET list-mode data and performed reconstruction using the ordered-subsets expectation maximization (OSEM) algorithm with 2 iterations and 21 subsets without AC. On the other hand, to generate the AC full-count PET image (i.e. AC-SDPET), we used the original PET list-mode data without down-sampling and performed the same OSEM protocol but with AC. For both reconstructions, a post-reconstruction Gaussian filter with $5 mm$ full width at half maximum (FWHM) was used. The voxel size of the reconstructed image was $4.08\times 4.08 \times 4.06 mm^3$. The image size was $200 \times 200$ in the transverse plane and varied in the axial direction depending on the patient's height. The 147 subjects were split into 120 subjects for training and 27 subjects for evaluation. We trained and implemented all models on the NVIDIA H100 GPUs. We trained our MADM model with a batch size of 50 for 200K training step with the $lr = 1 \times 10^{-4}$ and linearly decrease to 0. We used the EMA rate of 0.9999. We used the linear schedule of 1000 time steps. 



\subsection{Evaluation Strategies and Baselines}
Using the AC-SDPET as the reference, we evaluated the 3D translation performance both quantitatively and qualitatively. First, the quality of the translated images was evaluated at the image level using three image-quality metrics, including Peak Signal-to-Noise Ratio (PSNR), Structural Similarity Index (SSIM), and Root Mean Square Error (RMSE). SSIM focuses on the evaluation of structural recovery, while RMSE with a unit of SUV and PSNR with a unit of dB stress the evaluation of intensity profile recovery. Second, the quality of the translated PET was evaluated at clinically important local lesion regions throughout the whole body. Specifically, the difference in tumors' SUV mean was analyzed. For comparative evaluation, we compared our results against previous state-of-the-art DL-based image translation methods, including UNet-based methods \cite{ronneberger2015u}, GAN-based methods \cite{isola2017image}, and Diffusion-based methods \cite{lee2023improving,bieder2023diffusion}. For both quantitative and qualitative evaluations, we evaluated the performance under two different ultra-low-dose/count settings, i.e. 10\% and 5\% of PET full counts.  

\section{Experimental Results}
In Figure \ref{fig:comp_methods}, we show the qualitative comparison between our method and previous state-of-the-art methods for a patient example under the $5\%$ low-count PET scenario. As we can see, the original NAC-LDPET not only suffers from high noise levels but also highly biased quantification errors due to the attenuation effect, thus resulting in a low PSNR of $28.531$ dB. The error map computing the difference between it and the AC-SDPET, i.e. ground truth, further demonstrates this issue. Using previous CNN methods, i.e. UNet \cite{10.1007/978-3-319-24574-4_28} and GAN \cite{isola2017image}, we can see that these methods can generate reasonable AC-SDPET with PSNR reach to $36.73$ dB. However, relatively high errors were found in the liver and rib regions. On the other hand, we can see that previous diffusion methods, i.e. 3D-DDPM \cite{bieder2023diffusion} and TPDM\cite{lee2023improving}, can also generate reasonable AC-SDPET from NAC-LDPET. However, the performance in terms of PSNR and RMSE is slightly lower than the CNN-based methods. In the right column, our MADM that uses the CNN-based method's generation as prior and further refined using a multi-view 2.5D diffusion model generates the most consistent AC-SDPET as compared to ground truth, with further reduced error in major organs, e.g. liver and bone. 

In Figure \ref{fig:comp_methods_10NAC}, we present another example of AC-SDPET generated from $10\%$ low-count PET, comparing our method with previous methods. This example focuses on the tumor regions, which are zoomed in and highlighted with an orange arrow in Figure \ref{fig:comp_methods_10NAC}. Compared to other CNN-based and diffusion-based methods, our MADM not only achieves higher PSNR and lower NMSE quantitatively, but also shows a smaller error in the tumor region, as evidenced by the error map. Although both state-of-the-art methods and our MADM exhibit larger errors in the tumor region, our MADM significantly reduces errors in the surrounding areas. This reduction in error provides a substantial benefit for lesion masking and further analysis.

The quantitative comparisons were summarized in Table \ref{tab:comp_methods}. Similar to the observations from the visualizations, we can see that the traditional CNN-based approaches can slightly outperform the diffusion-based approach in our 3D scenario. For example, the cGAN\cite{Isola_2017_CVPR} achieved the best performance among all the baselines with PSNR = $38.295$ dB under the $5\%$ low-count scenario, while the best previous diffusion method, i.e. 3D-DDPM\cite{bieder2023diffusion}, only achieved PSNR = $37.197$ dB. Our method that combines the CNN and Diffusion achieved consistently better results with PSNR = $38.920$ dB than all the previous methods. 

\begin{table} [htb!]
\footnotesize
\centering
\caption{Quantitative comparison of AC-SDPET generated from NAC-LDPET data. Best results are marked in \textbf{bold}.}

\label{tab:comp_lesion}
\resizebox{0.48\textwidth}{!}{
    \begin{tabular}{l|c|c}
        \hline
        \multirow{1}{*}{\textbf{Lesion}}     & \multicolumn{2}{c}{\textbf{Error - SUV mean}}       \Tstrut\Bstrut\\
        \cline{2-3}
        \textbf{Quantification}              & 5\% Low-count            & 10\% Low-count        \Tstrut\Bstrut\\
        \hline
        NAC-LD                               & $0.1540 \pm 0.0710$   & $0.1517 \pm 0.0740$     \Tstrut\Bstrut\\
        \hline
        UNet \cite{10.1007/978-3-319-24574-4_28}                       & $0.0989 \pm 0.0394$   & $0.0855 \pm 0.0451$     \Tstrut\Bstrut\\
        \hline
        cGAN \cite{Isola_2017_CVPR}                         & $0.0894 \pm 0.0356$    & $0.0805 \pm 0.0393$     \Tstrut\Bstrut\\
        \hline
        3D-DDPM \cite{bieder2023diffusion}                      & $0.0907 \pm 0.0363$   & $0.0874 \pm 0.0368$     \Tstrut\Bstrut\\
        \hline
        TPDM \cite{lee2023improving}                         & $0.0922 \pm 0.0359$   & $0.0911 \pm 0.0399$     \Tstrut\Bstrut\\
        \hline
        MADM                                 & $\mathbf{0.0787 \pm 0.0391}$   & $\mathbf{0.0736 \pm 0.0403}$   \Tstrut\Bstrut\\
        \hline
    \end{tabular}
}
\end{table}

The quantitative comparisons of local tumor regions were summarized in Table \ref{tab:comp_lesion}. Similar to the observations from the above global quality analysis, we found that NAC-LDPET suffers from significant lesion quantification errors thus resulting in an averaged SUV mean error over $0.15$ in terms of RMSE under the $5\%$ condition. Even though previous methods can significantly reduce lesion quantification errors, the performance reaches its limit at approximately $0.09$. On the other hand, our MADM can further reduce this error to below $0.08$ which significantly improves the lesion quantification.  

\subsection{Ablative Studies}
\textbf{Multi-view Averaging Strategies} in MADM was studied and summarized in Table \ref{tab:branches}. Specifically, we investigated MADM's performance when different single view or combination of views strategies were implemented in MADM. As we can see, single-view strategies (first three rows) show consistently lower performance than the two-view combination strategies (fourth to seventh rows). MADM with all three views, i.e. axial, coronal, and sagittal, yields superior performance as compared to MADM with only one or two views.

\begin{table} [htb!]
\footnotesize
\centering
\caption{Ablative studies on the inclusion of multi-view averaging strategies in MADM. \textcolor{green}{\cmark} and \xmark\space means used and not used the model of the view in MADM, respectively. 5\% NAC-LDPET is considered here.}
\label{tab:branches}
\begin{tabular}{|c c c ||c|c|c|}
    \hline
    Axial                   & Coronal                & Saggital               & PSNR                   & RMSE               \Tstrut\Bstrut\\
    \hline   
    \textcolor{green}{\cmark}    & \xmark                      & \xmark                      & $37.884 \pm 1.460$         & $0.0378 \pm 0.0104$    \Tstrut\Bstrut\\
    \xmark                       & \textcolor{green}{\cmark}   & \xmark                      & $37.801 \pm 1.459$         & $0.0385 \pm 0.0098$    \Tstrut\Bstrut\\
    \xmark                       & \xmark                      & \textcolor{green}{\cmark}   & $38.066 \pm 1.458$         & $0.0362 \pm 0.0095$    \Tstrut\Bstrut\\
    \textcolor{green}{\cmark}    & \textcolor{green}{\cmark}   & \xmark                      & $38.350 \pm 1.505$         & $0.0341 \pm 0.0097$    \Tstrut\Bstrut\\
    \xmark                       & \textcolor{green}{\cmark}   & \textcolor{green}{\cmark}   & $38.465 \pm 1.468$         & $0.0331 \pm 0.0089$    \Tstrut\Bstrut\\
    \textcolor{green}{\cmark}    & \xmark                      & \textcolor{green}{\cmark}   & $38.747 \pm 1.486$         & $0.0311 \pm 0.0091$    \Tstrut\Bstrut\\
    \textcolor{green}{\cmark}    & \textcolor{green}{\cmark}   & \textcolor{green}{\cmark}   & $\mathbf{38.920 \pm 1.503}$   & $\mathbf{0.0299 \pm 0.0088}$    \Tstrut\Bstrut\\
    \hline
\end{tabular}
\end{table}

\textbf{Multi-view Averaging vs Sequential-based Strategies} was investigated and reported in Table \ref{tab:comp_sequential}. Specifically, we investigated a variant of MADM, called 2.5D Multi-View Sequential Diffusion (MVSD) that uses the same sets of 2.5D diffusion models as in MADM but was applied sequentially at multiple views during the sampling process. As we can see from the table, the MVSD approach, regardless of the order of axial, coronal, and sagittal views processed, exhibits negligible differences in performance, with PSNR values hovering around $38.61$ dB and RMSE at approximately $0.0321$. Our MADM on the last row achieved superior performance than all MVSD with different settings.

\begin{table} [htb!]
\footnotesize
\centering
\caption{Quantitative comparison of AC-SDPET generated from NAC-LDPET data. Best results are marked in \textbf{bold}. 5\% NAC-LDPET is considered here.}
\label{tab:comp_sequential}
    \begin{tabular}{l|c|c}
        \hline
        2.5D Multi-View Sequential                  & PSNR            & RMSE        \Tstrut\Bstrut\\
        \hline
        Axial $\rightarrow$ Coronal $\rightarrow$ Saggital    & $38.613 \pm 1.500$   & $0.0321 \pm 0.0092$     \Tstrut\Bstrut\\
        \hline
        Coronal $\rightarrow$ Saggital $\rightarrow$ Axial    & $38.609 \pm 1.500$   & $0.0321 \pm 0.0092$     \Tstrut\Bstrut\\
        \hline
        Saggital $\rightarrow$ Axial $\rightarrow$ Coronal    & $38.613 \pm 1.499$   & $0.0321 \pm 0.0092$     \Tstrut\Bstrut\\
        \hline
        \hline
        MADM                                 & $\mathbf{38.920 \pm 1.503}$   & $\mathbf{0.0299 \pm 0.0088}$   \Tstrut\Bstrut\\
        \hline
    \end{tabular}
\end{table}

\begin{table} [htb!]
\footnotesize
\centering
\caption{Quantitative comparison of AC-SDPET generated from NAC-LDPET data. Best results are marked in \textbf{bold}. 5\% NAC-LDPET is considered here.}
\label{tab:comp_2d_2_5d_diffs}
    \begin{tabular}{l|c|c}
        \hline
        Eval                 & PSNR            & RMSE        \Tstrut\Bstrut\\
        \hline
        2D MADM    & $37.816 \pm 1.541$   & $0.0386 \pm 0.0110$     \Tstrut\Bstrut\\
        \hline
        2.5D MADM s = 4   & $38.759 \pm 1.450$   & $0.0309 \pm 0.0080$     \Tstrut\Bstrut\\

        \hline
        2.5D MADM s = 8  & $\mathbf{38.920 \pm 1.503}$   & $\mathbf{0.0299 \pm 0.0088}$   \Tstrut\Bstrut\\
        \hline
        2.5D MADM s = 12   & $38.772 \pm 1.478$   & $0.0309 \pm 0.0087$     \Tstrut\Bstrut\\
        \hline
    \end{tabular}
\end{table}

\begin{figure}[htb!]
\centering
\includegraphics[width=0.445\textwidth]{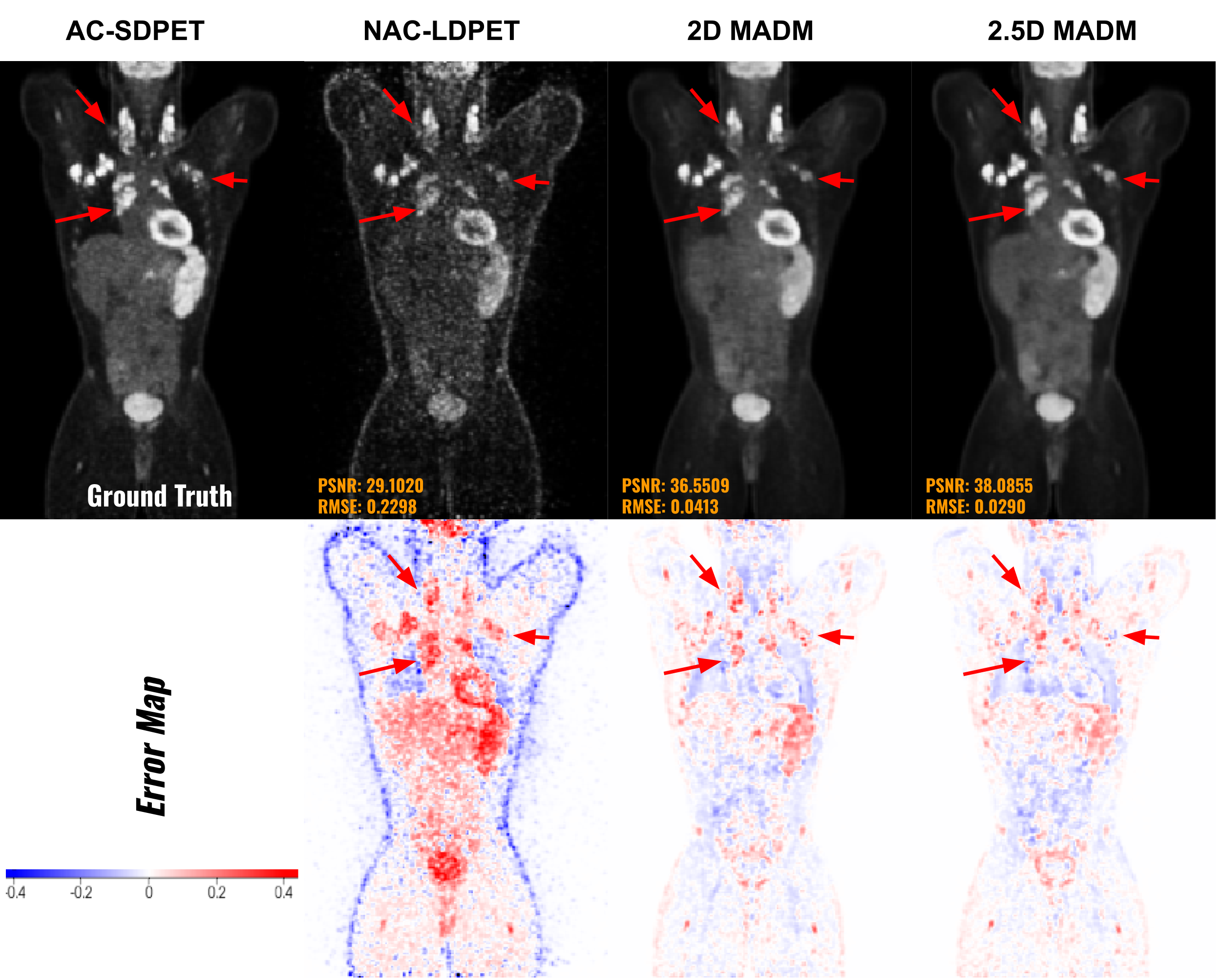}
\caption{Visual comparison of AC-SDPET generation form 2D MADM and 2.5D MADM under 5\% NAC-LDPET settings. The coronal view image(top) and error map(bottom) are shown. RMSE and PSNR values are calculated for each volume. The CT-based AC-SDPET(top) and image index into the color map of the error map(bottom) are shown in the first column.}
\label{fig:comp_2D_2.5D}
\end{figure}

\textbf{Impact of 2.5D Setting in MADM} was investigated and summarized in Table \ref{tab:comp_2d_2_5d_diffs}. As we can see, transitioning from a 2D to a 2.5D in MADM elevates the PSNR from $37.82$ dB to $38.92$ dB and reduces the RMSE from $0.0386$ to $0.0299$. This enhancement in model performance is attributed to the 2.5D model's capability to integrate contextual information from adjacent slices. Figure \ref{fig:comp_2D_2.5D} exemplifies that the 2.5D MADM more accurately predicts the shape and intensity in lesion areas (red arrows). By incorporating $s$ upper and lower slices relative to the target slice into its input, the 2.5D model achieves a more comprehensive understanding and precise reconstruction of abnormal areas. We found that setting the number of adjacent slices to $s = 8$ yields the best performance, achieving a PSNR of $38.92$ dB and an RMSE of $0.0299$.

\begin{figure}[htb!]
\centering
\includegraphics[width=0.46\textwidth]{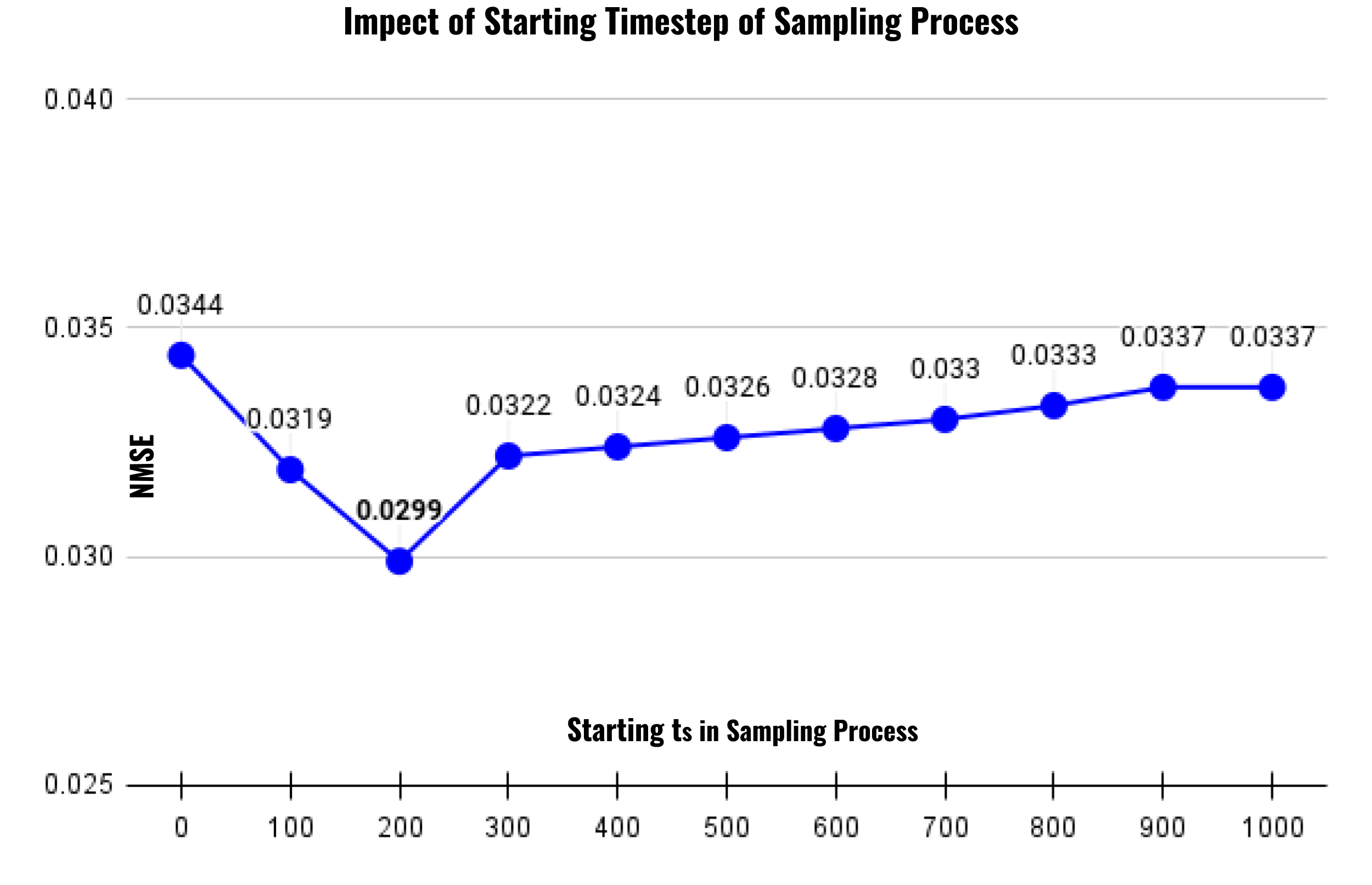}
\caption{Ablative study on the starting time step of the diffusion sampling process. NMSE evaluation under the $5\%$ NAC-LDPET setting is used here. Best performance is reached when $t_s=200$.}
\label{fig:line_plot_ts}
\end{figure}

\begin{figure}[htb!]
\centering
\includegraphics[width=0.455\textwidth]{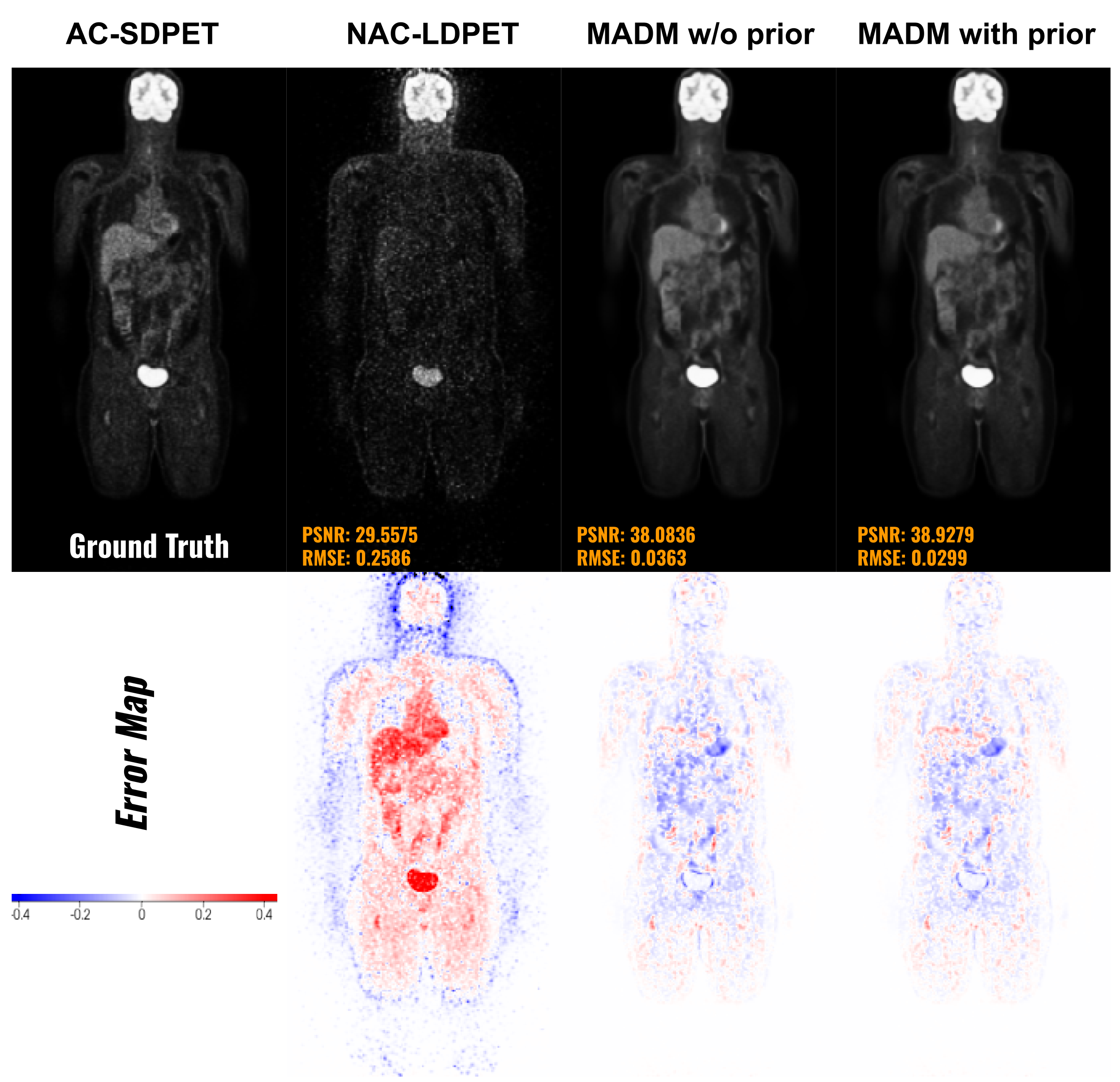}
\caption{Visual comparison of AC-SDPET generation with and without UNet Prior in MADM. under 5\% NAC-LDPET settings. The coronal view image(top) and error map(bottom) are shown. RMSE and PSNR values are calculated for each individual volume. The CT-based AC-SDPET(top) and image index in to the color map of the error map(bottom) are shown in the first column.}
\label{fig:plot_prior}
\end{figure}

\textbf{Impact of CNN-Prior of MADM} was investigated as well. A visual example is shown in Figure \ref{fig:plot_prior}. As we can see, implementing a prior strategy within the MADM framework significantly enhances the accuracy of the translated 3D image and reduces the intensity errors here. In parallel, we also studied the impact of starting timestep $t_s$ of using the prior image in the sampling process, and the results are summarized in Figure \ref{fig:line_plot_ts}, where we plotted the translation's NMSE against $t_s$ at an interval of $100$. As we observe from this result, the NMSE first decreases as $t_s$ increases. It reached the minimal NMSE of $0.0299$ at $t_s=200$ before the NMSE started to increase again. In contrast to previous diffusion methods that typically require $t_s=1000$, our method with optimal performance at $t_s=200$ outperforms them in terms of both speed and image quality. Please note that the result reported at $t_s = 0$ is c-GAN \cite{Isola_2017_CVPR} without diffusion process.


\section{Discussion} 
In this work, we develop a novel 3D image-to-image translation diffusion model, called 2.5D Multi-view Averaging Diffusion Model (MADM), for reducing the radiation in PET acquisition by directly translating NAC-LDPET to AC-SDPET in 3D. During the training, MADM trains multiple lightweight 2.5D models from axial, coronal, and sagittal views. In parallel, a conventional 3D CNN-based network (i.e. 3D UNet here) is also trained. During the inference process, we first use the trained 3D UNet for generating a prior 3D image. Then, we utilize this 3D prior image with scheduled noise as a starting point in our diffusion process to further refine it. During the subsequent diffusion, all views’ 2.5D models are applied in each sampling step and ensemble before proceeding to the next step. There are several key advantages of MADM. First, directly extending the previous 2D diffusion model into 3D is highly infeasible due to the challenge of large GPU memory requirements. MADM utilizes multiple lightweight 2.5D diffusion models instead of a 3D diffusion model, which makes it trainable on limited GPU memory settings, especially when dealing with high-resolution 3D image-to-image generation, e.g. NAC-LDPET to AC-SDPET 3D translation that explored in this work. As we can observe from the last two columns in Table \ref{tab:comp_methods}, MADM training memory consumption (per batch) is nearly 10 times less than 3D-DDPM. Second, MADM uses two simple and efficient strategies to enable high-quality 3D translation. The first is to use a 2.5D model that can learn the translation with contextual information from adjacent slices. The second is to use 2.5D models from three different views, where we average their outputs during the sampling process to ensure the slice-by-slice inconsistencies are minimized thus improving the 3D translation. Third, training diffusion models often require large-scale datasets to achieve robust performance. In contrast to directly feeding limited amounts of whole 3D volumes into a 3D diffusion model for training, 2.5D inputs from multiple views extracted from 3D volumes naturally improve the number and diversity of the training data. In fact, as we can observe from Table \ref{tab:comp_methods} and Table \ref{tab:comp_lesion}, MADM achieved significantly better translation in terms of both global and local metrics, and we believe this is one of the main reasons. Lastly, we utilize the one-step inference 3D CNN model to generate a prior image as the starting point of the diffusion model, which not only reduces the number of required sampling steps but improves the translation accuracy. 


However, our current work also has potential limitations, suggesting interesting directions for future studies. First, even though we reduce the memory requirement (Table \ref{tab:comp_methods}) for diffusion-based 3D image-to-image translation, the inference speed of MADM per 3D volume in our applications takes about 10 minutes for the volume $192 \times 192 \times 192$ and up to 30 minutes for the volume $192 \times 192 \times 460$ which one would consider long, especially when comparing with CNN-based methods. Depending on the user's time budget, MADM can adjust the starting time point of the diffusion process (Figure \ref{fig:line_plot_ts}), trading the inference speed with the performance. As we can see from Figure \ref{fig:line_plot_ts}, the optimal performance is achieved when $t_s=200$. In this case, we can adjust the $t_s=100$ to reduce the inference steps by two times, thus significantly reducing the inference time. While the performance reduces from NMSE of $0.0299$ to $0.0319$, the performance is still better than all the previous baselines in Table \ref{tab:comp_methods}. In parallel, we can adjust the number of views of MADM to improve the inference speed. As we can see from Table \ref{tab:branches}, even though optimal performance was achieved when all three views were included, we can use the two-view setting, i.e. axial + saggital views, to reduce the inference time but still generate superior 3D translation performance as compared to previous baseline methods (Table \ref{tab:comp_methods}). On the other hand, we could also replace the current 3D prior image generation network, i.e. 3D UNet, with more advanced network architecture, such as Transformer \cite{vaswani2017attention,dalmaz2022resvit} and Multi-stage Network \cite{zamir2021multi,zhou2022dudoufnet} to improve the prior 3D image generation, which may shorten the diffusion sampling process for optimal performance. Second, even though our results (Table \ref{tab:comp_methods}) show that MADM achieved superior 3D translation performance as compared to previous state-of-the-art diffusion and CNN-based methods, one of the key limitations is that we don't know how confident we are with our translation. 3D pixel-wise uncertainty estimation is highly desirable. In fact, uncertainty estimation for the I2I diffusion model is already well-established in many previous works \cite{zhou2024cascaded,gong2024pet}. In general, pixel-wise uncertainty can be obtained by running the diffusion reverse pass multiple times and then computing the pixel-wise standard deviation of the translated results. Similar here for MADM, we could also run multiple reverse passes by adding different noise to the same prior image, thus generating multiple 3D translation results. Then, the 3D pixel-wise uncertainty can be obtained by calculating the standard deviation of the translations. With the uncertainty map, one can identify areas where the model is confident and areas where it is not. One could use this information to fine-tune the model to reinforce the learning in the uncertain regions, thus potentially further improving the performance. In parallel to uncertainty estimation, averaging multiple runs' transition results as final translation also presents potential opportunities for further improving the performance \cite{zhou2024cascaded}. However, please note that utilizing this simple and efficient strategy for uncertainty estimation and performance gain comes at the cost of further elevated inference time, where the MADM inference time would increase linearly as the number of reverse runs increases. We recommend that users balance this trade-off in their application, especially in time-sensitive applications. Third, we only validated our 3D translation method for the application of PET dose reduction in this work. While we validate our method on clinically significant regions, i.e. tumors (Table \ref{tab:comp_lesion}), further analysis on the impacts on downstream analysis, such as prognosis and treatment planning, needs to be carried out in the future. In addition, interesting future directions also include validating MADM to other 3D medical imaging datasets and applications, such as 3D cross-modal MRI synthesis \cite{li2023brain,ji2022synthetic}, 3D contrast image synthesis in CT/MRI \cite{li2022multi,choi2021generating,liu2020dyefreenet}, Low-dose CT denoising \cite{yin2019domain,zhou2021dudodr,zhou2022dudoufnet}, and accelerated MRI reconstruction \cite{zhou2023recent,zhou2020dudornet,zhou2022dsformer}. Finally, we would also like to add that - while we framed the MADM as an independent method for 3D translation here, we believe this 2.5D-based method could be used for assisting the 3D network-based diffusion translation once the computation barrier for it is not an issue in the future. Specifically, during the diffusion inference process, we could average the 3D results with the multi-view 2.5D results, and it could potentially further improve the translation performance. This is an interesting direction that we will investigate in our future works. 

\section{Conclusion} 
In this work, we developed a novel 2.5D Multi-view Averaging Diffusion Model (MADM) for 3D medical image translation with the first application on low-count PET reconstruction with CT-less AC. Our method has three key components. First, to reduce the computational demands associated with 3D models and to mitigate the discontinuities inherent in 2D models, we proposed to use a 2.5D model that incorporates several adjacent slices in addition to the target slice for prediction. Second, to further minimize discontinuities and enhance prediction accuracy, we proposed to train separate 2.5D models for each view in 3D. During the inference, we average the result in each denoising step of the diffusion process. Third, to accelerate the translation process and refine generated image accuracy, we proposed to utilize 3D predictions from a CNN-based model as a prior for our diffusion model. Our experimental results show that MADM can generate high-quality AC standard-count PET(AC-SDPET) directly from the NAC low-count PET(NAC-LDPET) with limited computation memory resources, and outperforming several previous baselines. We believe our method can be extended to other 3D imaging applications, and potentially useful in other clinical applications. 

\section*{Acknowledgments}
This work was supported by the National Institutes of Health (NIH) grant R01EB025468 and R01CA275188.

\section*{Declaration of Competing Interest}
The authors declare that they have no known competing financial interests or personal relationships that could have appeared to influence the work reported in this paper.

\section*{Credit authorship contribution statement }
\textbf{Tianqi Chen}: Conceptualization, Methodology, Software, Visualization, Validation, Formal analysis, Writing original draft.
\textbf{Jun Hou}: Conceptualization, Methodology, Software, Visualization, Validation, Formal analysis, Writing - review and editing.
\textbf{Yinchi Zhou}: Results analysis, Writing - review and editing.
\textbf{Huidong Xie}: Conceptualization, Methodology, Software, Writing - review and editing.
\textbf{Xiongchao Chen}: Writing - review and editing.
\textbf{Qiong Liu}: Writing - review and editing.
\textbf{Xueqi Guo}: Writing - review and editing.
\textbf{Menghua Xia}: Writing - review and editing.
\textbf{James S. Duncan}: Writing - review and editing
\textbf{Chi Liu}: Writing - review and editing, Co-supervision.
\textbf{Bo Zhou}: Conceptualization, Methodology, Software, Visualization, Validation, Formal analysis, Writing original draft, Supervision.

\bibliographystyle{model2-names.bst}\biboptions{authoryear}
\bibliography{refs}

\end{document}